\documentclass[conference]{IEEEtran}
\IEEEoverridecommandlockouts
\usepackage{cite}
\usepackage{amsmath,amssymb,amsfonts}
\usepackage{algorithmic}
\usepackage{graphicx}
\usepackage{textcomp}
\usepackage{xcolor}
\usepackage{subfig}
\usepackage{caption}
\usepackage{booktabs}
\def\BibTeX{{\rm B\kern-.05em{\sc i\kern-.025em b}\kern-.08em
    T\kern-.1667em\lower.7ex\hbox{E}\kern-.125emX}}
\begin{document}

\title{A Lightweight Model-Driven 4D Radar Framework for Pervasive Human Detection in Harsh Conditions\\
}

\author{
\IEEEauthorblockN{Zhenan Liu}
\IEEEauthorblockA{\textit{Mechanical \& Mechatronics Engineering} \\
\textit{University of Waterloo}\\
Waterloo, Canada \\
z634liu@uwaterloo.ca}
\and
\IEEEauthorblockN{Amir Khajepour}
\IEEEauthorblockA{\textit{Mechanical \& Mechatronics Engineering} \\
\textit{University of Waterloo}\\
Waterloo, Canada \\
a.khajepour@uwaterloo.ca}
\and
\IEEEauthorblockN{George Shaker}
\IEEEauthorblockA{\textit{Electrical \& Computer Engineering} \\
\textit{University of Waterloo}\\
Waterloo, Canada \\
gshaker@uwaterloo.ca}
}

\maketitle


\begin{abstract}
Pervasive sensing in industrial and underground environments is severely constrained by airborne dust, smoke, confined geometry, and metallic structures, which rapidly degrade optical and LiDAR-based perception. Elevation-resolved 4D millimeter-wave (mmWave) radar offers strong resilience to such conditions, yet there remains a limited understanding of how to process its sparse and anisotropic point clouds for reliable human detection in enclosed, visibility-degraded spaces. This paper presents a fully model-driven 4D radar perception framework designed for real-time execution on embedded edge hardware. The system uses radar as its sole perception modality and integrates domain-aware multi-threshold filtering, ego-motion–compensated temporal accumulation, KD-tree Euclidean clustering with Doppler-aware refinement, and a rule-based 3D classifier. The framework is evaluated in a dust-filled enclosed trailer and in real underground mining tunnels, and in the tested scenarios the radar-based detector maintains stable pedestrian identification as camera and LiDAR modalities fail under severe visibility degradation. These results suggest that the proposed model-driven approach provides robust, interpretable, and computationally efficient perception for safety-critical applications in harsh industrial and subterranean environments.
\end{abstract}


\section{Introduction}

Pervasive sensing in industrial, construction, and underground environments is fundamentally constrained by visibility. Airborne dust, smoke, confined geometry, and metallic infrastructure often cause optical sensors and LiDAR to fail, resulting in missed detections and unreliable situational awareness. Millimeter-wave (mmWave) radar, in contrast, is inherently resilient to illumination changes and particulate interference, making it a promising modality for such harsh environments.

Conventional automotive radars, however, provide only coarse angular resolution and typically lack elevation estimation, yielding sparse range--azimuth--Doppler measurements that are difficult to interpret in enclosed metallic spaces where multipath and ghost reflections are prevalent. Recent advances in \emph{4D imaging radar} introduce elevation estimation through large virtual MIMO arrays, producing 3D point clouds enriched with Doppler and radar cross section (RCS). Although these sensors have shown promise in outdoor visibility-degraded settings, there is no established consensus on how to process their anisotropic, noise-prone point clouds in confined industrial or underground environments dominated by dust and reflective surfaces.

\begin{figure}[t]
    \centering
    \includegraphics[width=\linewidth]{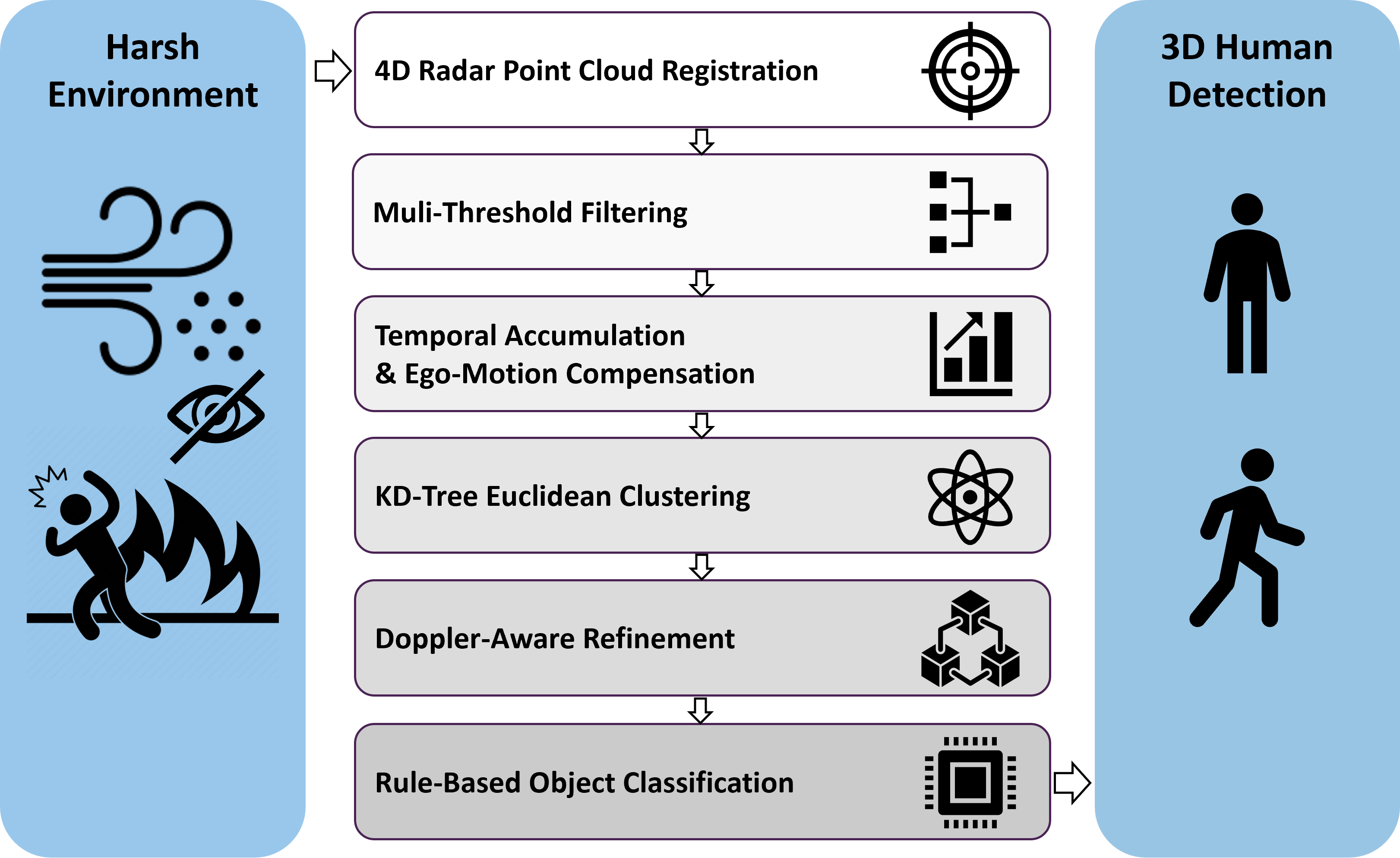}
    \caption{Overview of the model-driven 4D radar pipeline: raw radar measurements through multi-threshold filtering, ego-motion–compensated temporal accumulation, clustering, Doppler-aware refinement, rule-based 3D classification }
    \label{fig:pipeline_overview}
    \vspace{-0.4cm}
\end{figure}

From this context, three specific gaps emerge:

\begin{enumerate}
    \item \textbf{Sparse evaluation of 4D radar in enclosed environments with visibility degradation.}  
    Existing studies focus primarily on outdoor automotive scenes. Radar behaviour in dust-filled, metal-walled spaces, such as industrial corridors or underground tunnels, remains largely uncharacterized.

    \item \textbf{Limited development of radar-specific point-cloud processing pipelines.}  
    Compared with extensive research on optical and LiDAR perception in adverse weather, far fewer works investigate radar-specific filtering, Doppler conditioning, and cluster-level reasoning tailored to 4D radar noise characteristics.

    \item \textbf{Need for strong, interpretable model-driven baselines.}  
    The rapid emergence of learning-based radar detectors underscores the need for lightweight, physics-motivated baselines that operate reliably in harsh enclosed environments and provide stable targets for future learning-based extensions.
\end{enumerate}

To address these gaps, this paper proposes a \textbf{fully model-driven 4D radar perception framework} for real-time human detection in industrial and underground scenarios. The system uses 4D radar as its \emph{only perception modality}, while relying on external LiDAR--IMU odometry solely for ego-motion compensation and on LiDAR/camera data only for visualization and baseline comparison.

The main contributions of this work are:

\begin{itemize}
    \item \textbf{A radar-only, model-driven 4D perception pipeline} tailored for enclosed dusty and metallic environments, filling a gap left by prior work focused on outdoor driving, sensor fusion, or learning-based detectors.

    \item \textbf{A domain-aware multi-threshold filtering module} using RCS bounds, angular constraints, and Doppler plausibility to suppress multipath and clutter common in confined spaces.

    \item \textbf{A lightweight two-frame temporal accumulation scheme} that uses external LiDAR--IMU odometry solely for ego-motion compensation, improving spatial consistency while keeping the system real-time and embedded-friendly.

    \item \textbf{A Doppler-aware clustering and deterministic 3D classifier} that infers object type, motion state, and line-of-sight movement without training.

    \item \textbf{Experimental validation in controlled dust and underground tunnels}, demonstrating—\emph{in our tested scenarios}—that 4D radar maintains stable pedestrian detection where camera and LiDAR sensing are severely degraded.
\end{itemize}

\section{Related Work}

\subsection{Perception in Adverse Environments}

Optical and LiDAR-based perception degrades sharply under adverse visibility
conditions such as rain, fog, airborne dust, and low light. Numerous studies
report substantial losses in detection and tracking performance even with
advanced enhancement models~\cite{10733168,CHEN2026128994}. LiDAR suffers from
attenuation, backscatter, and spurious returns that reduce point-cloud
completeness~\cite{dreissig2023survey}. These limitations intensify in
confined or underground spaces where metallic infrastructure creates strong
multipath and where maintaining situational awareness is intrinsically
challenging~\cite{WANG2023104822}. These trends motivate sensing modalities
that remain reliable when optical pathways collapse.

\subsection{Conventional Automotive Radar}

Automotive mmWave radars (76--81\,GHz) provide robust range and Doppler
estimates and are widely deployed for highway safety~\cite{9348922,7870764}.
However, limited antenna aperture yields coarse angular resolution and no
direct elevation estimate. Standard radar outputs provide sparse
range--azimuth--Doppler detections rather than full 3D structure, and their
performance deteriorates in enclosed metallic environments due to multipath,
ghost reflections, and anisotropic lobes~\cite{9760734}. These constraints
limit the applicability of conventional radar to object-level perception in
industrial or underground spaces.

\subsection{4D Imaging Radar and Existing Datasets}

Recent 4D mmWave imaging radars provide elevation estimation through large
virtual MIMO arrays, generating anisotropic 3D point clouds enriched with
Doppler and RCS~\cite{han20234d}. This capability has inspired several
learning-based perception studies and datasets. TJ4DRadSet~\cite{9922539} and
View-of-Delft (VoD)~\cite{apalffy2022} provide radar, camera, and LiDAR data
for outdoor autonomous driving, including rain and artificial fog scenarios.
Other works examine radar–camera fusion or deep networks for pedestrian
detection~\cite{skog2024humandetection4dradar}.

However, these datasets and methods share two key limitations.  
First, their adverse-weather scenes are dominated by water-based fog or smoke
and occur in wide outdoor spaces. They do not capture the unique behavior of
mmWave radar in confined, dust-filled, metallic environments where cavity
propagation, strong multipath, and sparse returns differ fundamentally from
outdoor driving settings.  
Second, most perception pipelines rely on learning-based models or
radar–camera fusion, which require large labeled datasets and are difficult to
deploy on lightweight embedded hardware.

\subsection{Classical Geometric Pipelines}

Traditional geometric pipelines rely on nearest-neighbour search, clustering,
and principal-component analysis for bounding-box estimation~\cite{8443742}.
Euclidean clustering with KD-tree acceleration is widely used for LiDAR
processing~\cite{MIAO2023107951}, but such methods assume relatively dense and
isotropic point clouds. In contrast, 4D radar returns are sparse and anisotropic,
and their Doppler information is rarely exploited in classical frameworks.

\subsection{Novelty and Positioning of This Work}

Compared with prior research, this work differs in three fundamental ways:

\begin{enumerate}
    \item \textbf{Environment:} Prior 4D radar studies focus on outdoor
    autonomous driving, whereas this work targets enclosed, dust-filled,
    metallic environments representative of industrial and underground
    operations.

    \item \textbf{Modality and independence from vision:} Unlike radar–camera
    fusion or deep-learning detectors that require labeled datasets, our
    framework uses radar as the \emph{sole perception modality}, relying only
    on external odometry for motion compensation.

    \item \textbf{Model-driven, lightweight design:} Existing methods often
    employ heavy learning pipelines. We introduce a fully model-driven radar
    pipeline that integrates physically interpretable filtering, Doppler-aware
    clustering, and rule-based 3D classification for real-time performance on
    embedded hardware.
\end{enumerate}

To our knowledge, this is the first work to demonstrate a fully
model-driven, radar-only 4D perception framework operating in enclosed,
dust- and metal-filled industrial environments, without any reliance on optical
modalities or supervised learning.

\section{Methodology}

\subsection{4D Radar Platform and Multi-Sensor Setup}

The 4D mmWave imaging radar used in this work is a cascaded four-chip device equipped with 12 transmit (TX) and 16 receive (RX) antennas arranged into a large virtual MIMO array capable of resolving range, azimuth, elevation, and Doppler~\cite{altosradar2023}. Manufacturer specifications list azimuth and elevation resolutions on the order of $1$--$2^{\circ}$ and pedestrian detection ranges of approximately 200\,m; these values serve only as context, as all analysis relies on empirical measurements collected in our experiments.

Radar data are acquired using the vendor’s ROS driver, which receives UDP packets, decodes the raw range--angle--Doppler detections, converts them into Cartesian coordinates, and publishes the resulting elevation-resolved radar point cloud as \texttt{sensor\_msgs/PointCloud2} at 15\,Hz. Logging and timestamping are performed on an NVIDIA Jetson Orin NX edge computer. Extrinsic calibration between radar, LiDAR, and the IMU is maintained within the ROS \texttt{tf} tree (\texttt{radar}$\rightarrow$\texttt{base\_link}$\rightarrow$\texttt{world}), ensuring consistent geometric registration across all sensors.

The radar is rigidly co-mounted with a 40-line LiDAR and an infrared (IR) camera. LiDAR provides geometric reference and supplies LiDAR--IMU odometry used exclusively for ego-motion compensation; it does not participate in perception or decision-making. The IR camera documents visibility degradation and serves only as a qualitative baseline. All perception outputs in this work are derived solely from the 4D radar point clouds.

Each radar point is represented as
\begin{equation}
\mathbf{p}_i = (x_i, y_i, z_i, v_i^{\mathrm{dop}}, \mathrm{RCS}_i, \delta_i),
\end{equation}
where $(x_i,y_i,z_i)$ are Cartesian coordinates, $v_i^{\mathrm{dop}}$ is radial Doppler velocity, $\mathrm{RCS}_i$ is radar cross section, and $\delta_i$ is a vendor-supplied dynamic/static flag. Indoor radar frames typically contain 3\,000--6\,000 points.




\subsection{Domain-Aware Multi-Threshold Filtering}

The first processing stage removes spurious radar points arising from diffuse reflections, metallic multipath, and implausible Doppler measurements. Two threshold profiles are used:

\textit{Indoor profile:} strict angular and RCS limits to suppress wall- and ceiling-induced multipath in enclosed spaces (e.g., azimuth $[-5,5]^\circ$, elevation $[-2,8]^\circ$, RCS $[0,45]$).

\textit{Outdoor profile:} wider angular and RCS bounds to accommodate long-range reflections and broader operational coverage (e.g., azimuth $[-15,15]^\circ$, elevation $[-6,12]^\circ$, RCS $[-5,55]$).

Each raw radar point $\mathbf{p}_i$ is subjected to three filters.

\paragraph{1) RCS thresholding.}
Very weak diffuse returns and extremely strong multipath-induced peaks are rejected:
\begin{equation}
\mathrm{RCS}_{\min} < \mathrm{RCS}_i < \mathrm{RCS}_{\max}.
\end{equation}

\paragraph{2) Angular bound filtering.}
Azimuth $\theta_i$ and elevation $\phi_i$ must lie within the radar’s valid field of view:
\begin{equation}
\theta_i\in[\theta_{\min},\theta_{\max}],\qquad
\phi_i\in[\phi_{\min},\phi_{\max}].
\end{equation}
Points outside these bounds are frequently associated with ceiling/floor multipath or grazing-angle reflections.

\paragraph{3) Doppler plausibility filtering.}
Points with Doppler velocities inconsistent with human or machinery motion, or zero-Doppler points in locations not physically associated with static structures, are removed:
\begin{equation}
v_{\min} \le v_i^{\mathrm{dop}} \le v_{\max}.
\end{equation}

The resulting filtered set is
\begin{equation}
\small
\begin{aligned}
\mathcal{P}_{\mathrm{filtered}} =
\{\mathbf{p}_i\in\mathcal{P}_{\mathrm{raw}} \mid &
\mathrm{RCS}_{\min}<\mathrm{RCS}_i<\mathrm{RCS}_{\max},\\
&
\theta_i\in[\theta_{\min},\theta_{\max}],\;
\phi_i\in[\phi_{\min},\phi_{\max}],\\
&
v_i^{\mathrm{dop}}\in[v_{\min},v_{\max}]
\},
\end{aligned}
\end{equation}
executed in $\mathcal{O}(n)$ time per frame. This significantly reduces downstream fragmentation during clustering.

\subsection{Temporal Accumulation with Ego-Motion Compensation}

Radar point clouds are sparser than LiDAR and may contain only a few points per pedestrian in each frame. To improve spatial continuity while maintaining real-time performance, we accumulate \textbf{exactly two} consecutive frames.

Odometry is provided by FAST-LIO~\cite{xu2021fastlio2}, a tightly coupled LiDAR--IMU fusion system that outputs six-degree-of-freedom poses at high frequency on a ROS \texttt{/odom} topic. These poses are used solely to (i) compensate for ego-motion when aligning radar frames and (ii) compute the platform velocity $\mathbf{v}_{\mathrm{ego}}$ used in Doppler correction. Radar perception itself remains radar-only.

Let $\mathcal{P}_{t}$ and $\mathcal{P}_{t-1}$ denote the filtered radar point clouds at consecutive timestamps. The transform $\mathbf{T}_{t-1}^{t}\in SE(3)$ maps points from the $(t\!-\!1)$ frame to the $t$ frame. Because odometry updates run faster than radar frames, poses are time-synchronized and interpolated to radar timestamps.

The previous cloud is transformed as:
\begin{equation}
\mathcal{P}_{t-1}^{\,t}
=
\{\mathbf{T}_{t-1}^{t}\,\mathbf{p}_i \mid \mathbf{p}_i\in\mathcal{P}_{t-1}\},
\end{equation}
and the accumulated cloud becomes:
\begin{equation}
\mathcal{P}_{\mathrm{acc}}
=
\mathcal{P}_{t} \cup \mathcal{P}_{t-1}^{\,t}.
\end{equation}

The two-frame window improves spatial consistency without noticeable latency and without accumulating drift. Because the radar does not strongly sense the ground, no ground-plane removal or voxel downsampling is required.

\subsection{KD-Tree Euclidean Clustering and Doppler-Aware Filtering}

Let $\mathcal{P}_{\mathrm{acc}}=\{\mathbf{p}_i\}_{i=1}^N$. A KD-tree is constructed over the spatial coordinates $(x_i,y_i,z_i)$ to enable neighbour queries in $\mathcal{O}(\log N)$ time.

\paragraph{Euclidean clustering.}
Clusters $\{\mathcal{C}_1,\ldots,\mathcal{C}_K\}$ are generated via radius-based connected components:
\begin{equation}
\|\mathbf{p}_i-\mathbf{p}_j\|_2 < d_{\mathrm{th}}
\;\Rightarrow\;
\mathbf{p}_i,\mathbf{p}_j\in\mathcal{C}_k,
\end{equation}
with overall complexity $\mathcal{O}(N\log N)$. Although originally designed for dense, isotropic LiDAR, Euclidean clustering performs reliably here because (i) strict filtering removes multipath noise, (ii) temporal accumulation increases local density, and (iii) confined indoor geometry produces compact, spatially coherent radar returns. In our experiments, we set $d_{\mathrm{th}}=0.6m$ (comparable to half a human torso width at typical ranges) and require at least 3 points per cluster to avoid noise.

\paragraph{Cluster descriptors.}
For each cluster $\mathcal{C}_k$, we compute:

(1) Mean Doppler:
\begin{equation}
\bar{v}_k =
\frac{1}{|\mathcal{C}_k|}
\sum_{\mathbf{p}_i\in\mathcal{C}_k} v_i^{\mathrm{dop}}.
\end{equation}

(2) Ego-motion--compensated mean Doppler:
\[
\hat{\mathbf{r}}_i = \frac{(x_i,y_i,z_i)}{\|(x_i,y_i,z_i)\|_2},
\]
\begin{equation}
\bar{v}_k^{\mathrm{comp}} =
\frac{1}{|\mathcal{C}_k|}
\sum_{\mathbf{p}_i\in\mathcal{C}_k}
\left(v_i^{\mathrm{dop}}
- \mathbf{v}_{\mathrm{ego}}\cdot\hat{\mathbf{r}}_i\right).
\end{equation}

(3) Modal RCS:
\begin{equation}
\mathrm{RCS}_{\mathrm{mode},k}
=
\mathrm{mode}\{\mathrm{RCS}_i : \mathbf{p}_i\in\mathcal{C}_k\}.
\end{equation}

\paragraph{Cluster retention.}
A cluster is retained if
\begin{equation}
|\bar{v}_k^{\mathrm{comp}}| > v_{\min}
\quad\text{or}\quad
\mathrm{RCS}_{\mathrm{mode},k}
\in [\mathrm{RCS}_{\min},\mathrm{RCS}_{\max}].
\end{equation}
The Doppler criterion eliminates static structures, while the RCS criterion preserves both moving pedestrians and strong reflective objects.

\begin{figure}[t]
    \centering
    \includegraphics[width=0.95\linewidth]{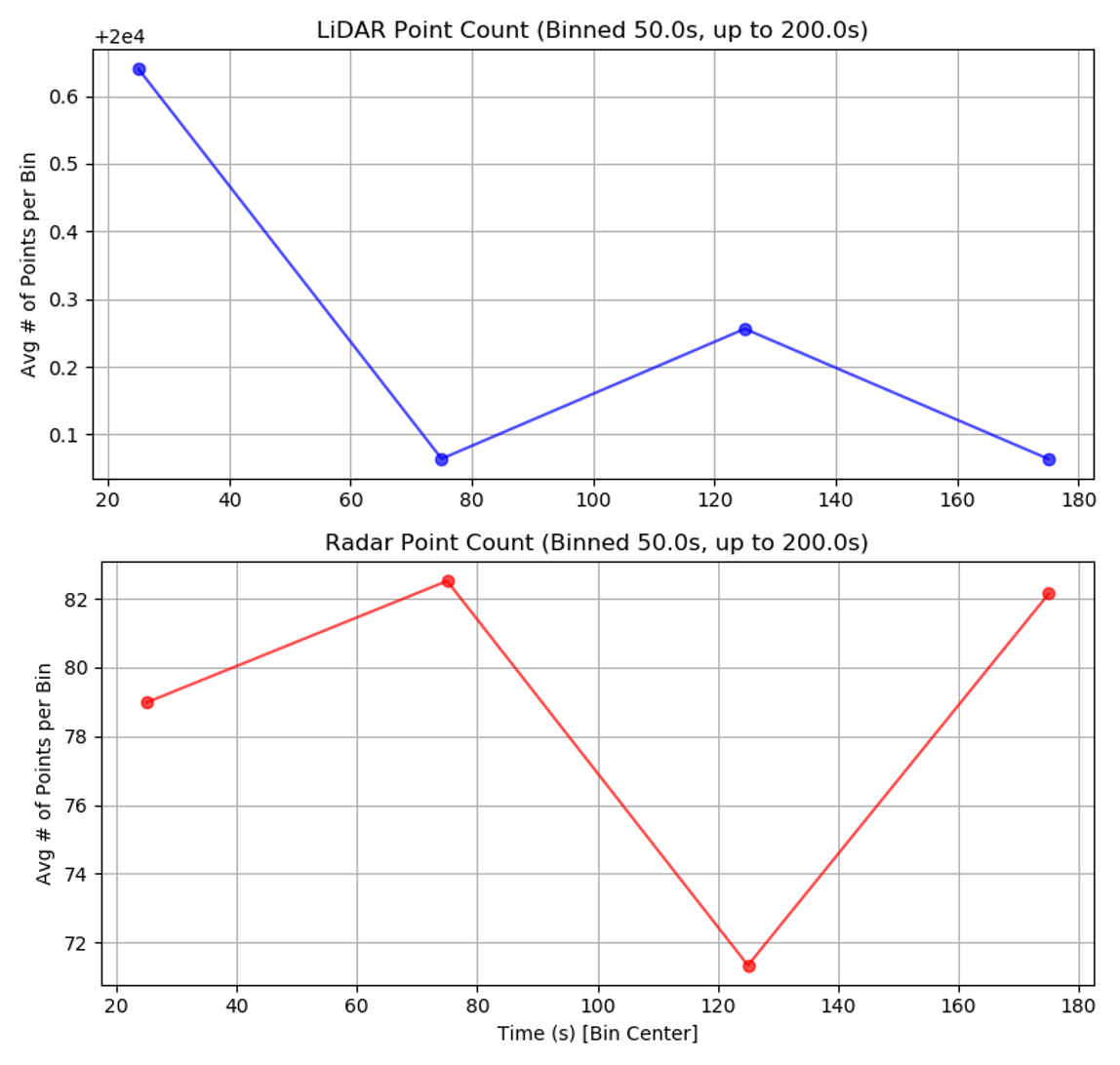}
    \caption{Average point count per sensor under increasing dust concentration. 
    LiDAR returns decline sharply due to attenuation and backscatter, while 4D radar maintains a nearly constant point count with only minor fluctuations, confirming robustness to particulate interference.}
    \label{fig:point_count}
    \vspace{-0.4cm}
\end{figure}

\subsection{Rule-Based 3D Object Classification}

Each retained cluster $\mathcal{C}_k^\ast$ is assigned semantic properties using three descriptors:

\paragraph{Type inference.}
Cluster dimensions $(w_k,l_k,h_k)$ are derived from its bounding box. Pedestrians typically exhibit widths of 0.5--1.0\,m, heights below 2\,m, and moderate RCS near 0; larger metallic objects exhibit both larger extents and higher RCS. Thresholds were tuned on a small validation subset and fixed for all experiments.

\paragraph{Motion state.}
The magnitude of $|\bar{v}_k^{\mathrm{comp}}|$ indicates whether the object is static or dynamic.

\paragraph{Heading (line-of-sight only).}
The sign of $\bar{v}_k^{\mathrm{comp}}$ indicates whether the object is approaching or receding along the radar line-of-sight. This is not a full 3D heading estimate but provides actionable situational awareness in confined environments.

The resulting classifier is deterministic, training-free, and runs in $\mathcal{O}(K)$ per frame, supporting real-time embedded deployment.

\begin{figure*}[h]
    \vspace{-0.2cm}
    \centering
    \subfloat[Clear environment (Dust Level~0).\label{fig:dust_level0}]{
        \includegraphics[width=0.47\linewidth]{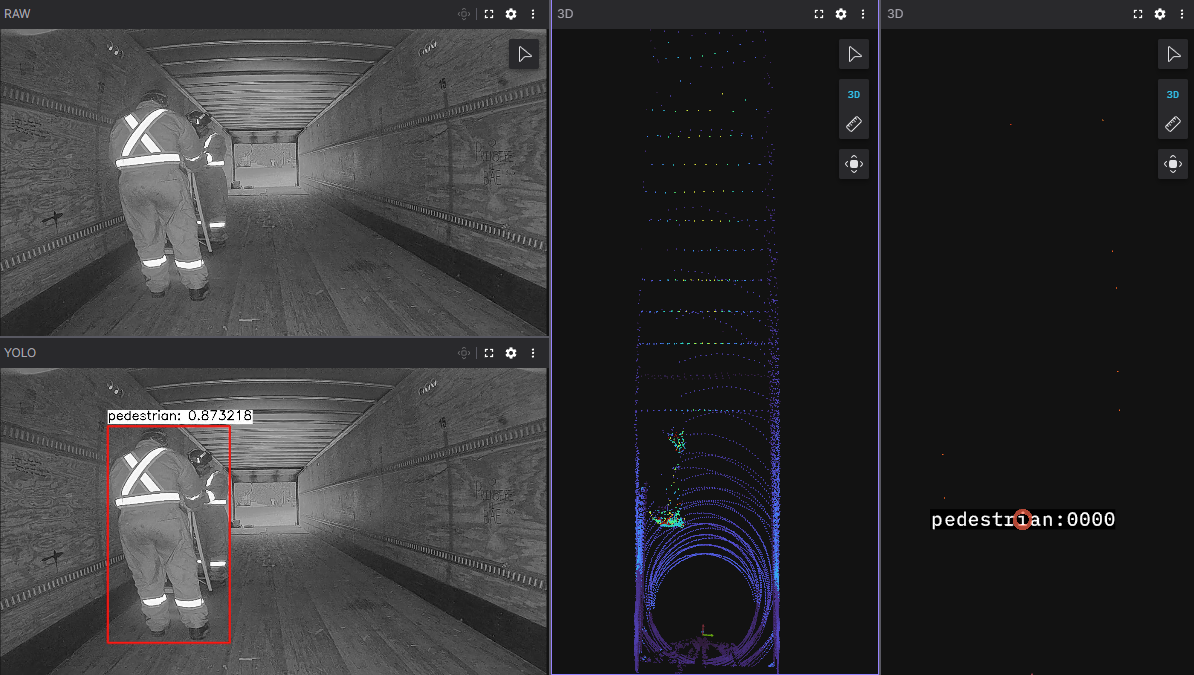}
    }
    \hfill
    \subfloat[Highest dust concentration.\label{fig:dust_level3}]{
        \includegraphics[width=0.47\linewidth]{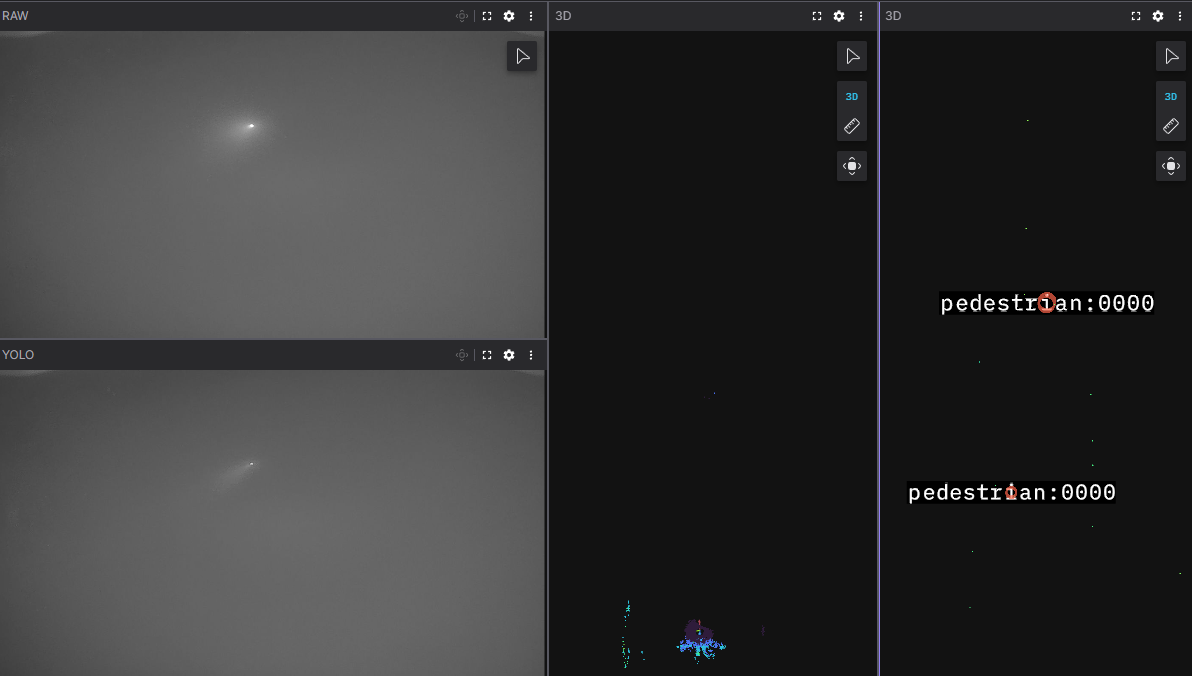}
    }
    \caption{Qualitative detection results under varying dust conditions. 
    Left: Under clear conditions, IR camera, LiDAR, and radar all provide consistent structure, 
    and both camera-based and radar-based detectors identify pedestrians. 
    Right: At the highest dust concentration, IR and LiDAR fail completely, 
    while the model-driven 4D radar framework continues to detect pedestrians robustly.}
    \label{fig:dust_levels_comparison}
    \vspace{-0.4cm}
\end{figure*}

\section{Experiments}

This section evaluates the proposed model-driven 4D radar perception framework
under two complementary settings: a controlled dust-filled environment enabling
repeatable stress testing, and an active underground mine demonstrating
real-world generalization. Evaluation focuses on (i) sensor-level robustness
under degraded visibility, (ii) object-level detection stability compared with
a vision-based detector, and (iii) qualitative performance in challenging
industrial scenarios. These metrics provide a meaningful proof of concept
without requiring dense 3D annotations, which are impractical for 4D radar due
to sparse, anisotropic scattering and ambiguous spatial extents.

\subsection{Controlled Dust-Filled Trailer Dataset}

Experiments were conducted inside a 53$'$~$\times$~10$'$~$\times$~11$'$ enclosed
trailer configured as a controlled test chamber. Wooden walls, ceiling beams,
and regularly spaced metal strips created structured but interpretable
reflective surfaces while avoiding excess diffuse multipath. The corridor-like
geometry emulates narrow industrial tunnels and underground passages.

A sand-based particulate mixture was dispersed to generate progressively
increasing visibility degradation, from clear conditions (Dust Level~0) to
heavy dust saturation. Two participants wearing high-visibility personal
protective equipment moved throughout the space, producing both dynamic and
near–zero-Doppler scenarios relevant for safety applications. Radar point
clouds, LiDAR scans, and infrared (IR) images were synchronized using ROS,
ensuring consistent timestamp alignment across modalities. No LiDAR or camera information is used for perception, classification, or decision
logic.

Figure~\ref{fig:point_count} summarizes the average number of returns per
sensor as dust concentration increases. LiDAR point density drops sharply due
to attenuation and backscatter, whereas the 4D radar maintains nearly constant
point density with only minor fluctuations. This confirms that the mmWave
modality remains structurally informative even under severe particulate
interference.

\paragraph*{Annotation considerations.}
The dataset does not include 3D bounding-box annotations. Radar measurements
exhibit sparse, anisotropic point distributions dominated by specular and
multipath reflections, preventing reliable spatial annotation. Constructing a
large fully annotated 4D radar dataset would require specialized tools and
substantial labor, which is outside the scope of this work. Instead, evaluation
emphasizes objective, interpretable metrics: sensor-level robustness,
high-level detection stability, and deployability on embedded hardware.

\subsection{Model-Driven Radar vs.\ Camera-Based Detection}

We compare the proposed radar-only classifier against a YOLOv8 camera-based
detector~\cite{Jocher_Ultralytics_YOLO_2023} pretrained on the Construction
Site Safety dataset~\cite{construction-site-safety_dataset}. Both methods
process synchronized ROS data streams and attempt to detect pedestrians as dust
levels increase.

\begin{figure}[!t]
    \centering
    \includegraphics[width=\linewidth]{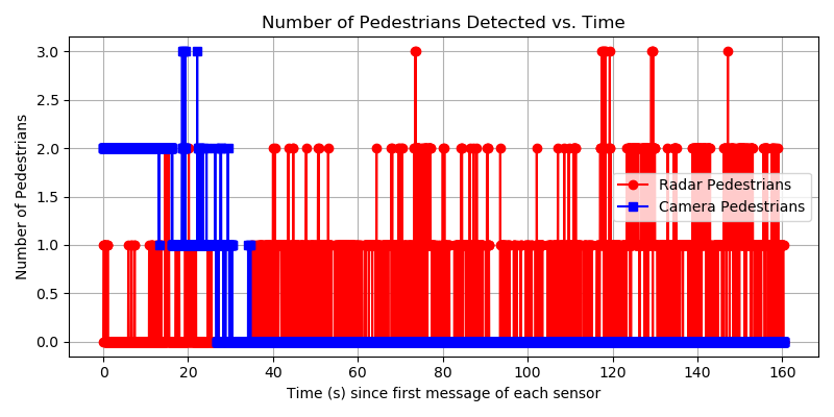}
    \caption{Pedestrian detections over time from the model-driven radar
    classifier (red) and a YOLOv8 camera detector (blue). As dust levels
    increase, camera-based detection collapses, whereas radar detection
    remains stable.}
    \label{fig:radar_camera_ped}
\end{figure}

Figure~\ref{fig:radar_camera_ped} shows the number of pedestrians detected over
time. In clear conditions, both detectors operate reliably. As dust
concentration rises, IR imagery becomes heavily obscured and YOLOv8 fails
abruptly. In contrast, the radar-based classifier remains stable across all
dust levels due to its reliance on mmWave reflections.

\paragraph*{Quantitative comparison.}

To complement the qualitative observations, we performed a lightweight but
objective statistical analysis using a short segment of synchronized radar and
camera data in which two pedestrians were continuously present across all
160 frames. The evaluation is conducted using \emph{count-based
metrics} that require no spatial labels due to the lack of annotations: (i) frame-wise recall, indicating
whether at least one pedestrian was detected in a given frame; (ii) person-count
recall, comparing the number of detected individuals against the known
two-person ground truth; and (iii) false-alarm rate, measuring frames in which
the detector overestimates the number of pedestrians. These metrics directly
capture temporal stability and robustness under dust without relying on
annotation-heavy pipelines.

\begin{table}[t]
\centering
\caption{Quantitative pedestrian-detection results in a two-person sequence (160 frames)}
\label{tab:quant_results}
\begin{tabular}{lccc}
\toprule
\textbf{Metric} & \textbf{Radar (Ours)} & \textbf{YOLOv8 (Camera)} \\
\midrule
Frame-wise recall 
& 150/160 (\textbf{94\%}) 
& 27/160 (17\%) \\
Person-count recall 
& 270/320 (\textbf{84\%}) 
& 40/320 (12.5\%) \\
False-alarm rate 
& $\sim$5\% of frames 
& $\sim$3\% of frames \\
\bottomrule
\end{tabular}
\vspace{-0.4cm}
\end{table}

These results show a clear performance gap: while the camera-based detector
fails in most frames due to dust-induced visibility loss, the radar-based
pipeline retains high recall and only minor count overestimation caused by
cluster fragmentation. The qualitative sequences in
Figs.~\ref{fig:dust_level0}--\ref{fig:dust_level3} reinforce this behavior:
under severe dust saturation, both the IR camera and LiDAR become unusable,
yet the 4D radar point cloud remains sufficiently structured for consistent and
reliable pedestrian detection.

\subsection{Underground Mining Scenario}

\begin{figure}[h]
    \centering
    \includegraphics[width=\linewidth]{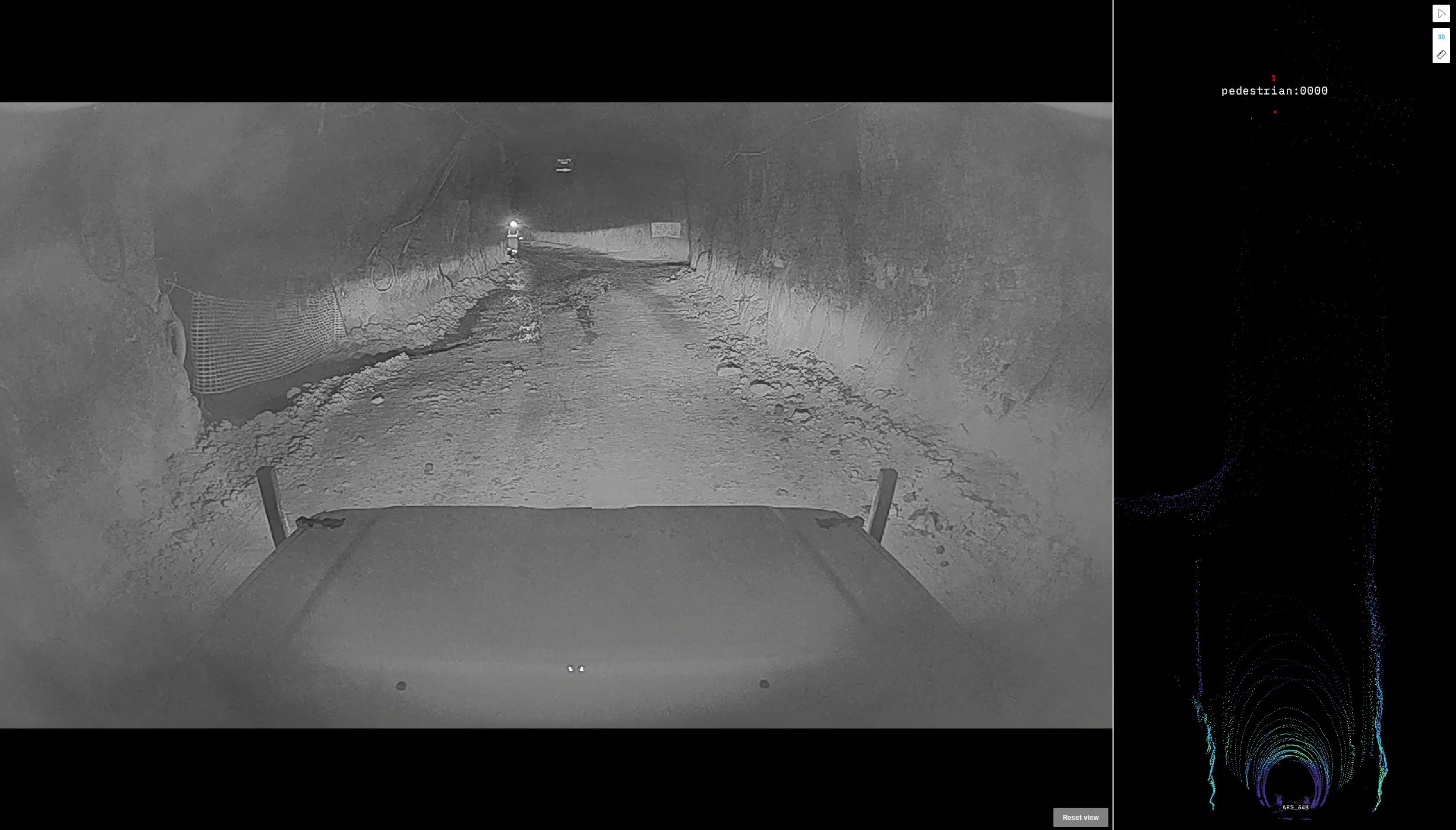}
    \caption{Underground mining evaluation. Camera-based detection fails due
    to dust and headlamp glare, whereas the model-driven radar framework
    maintains reliable pedestrian detection.}
    \label{fig:mining_scenario2}
    \vspace{-0.2cm}
\end{figure}

To evaluate generalization beyond controlled environments, the same framework was
deployed in an active underground mining tunnel:

\paragraph*{Scenario overview.}
Figure~\ref{fig:mining_scenario2} shows a representative frame from the underground deployment. A miner is present at approximately 60\,m distance, which makes the human signature appear small in both the IR camera view and the radar visualization. Airborne dust, irregular tunnel surfaces, and severe headlamp glare substantially degrade image contrast, causing no detections from YOLO. Despite the reduced visibility and long range, the model-driven 4D radar framework consistently identifies the pedestrian by leveraging Doppler- and RCS-aware clustering, demonstrating robustness where camera-based perception collapses.

Despite these conditions, the radar framework maintains stable detection across
frames. Doppler- and RCS-aware clustering isolates the pedestrian from
background multipath, demonstrating robustness in visibility-degraded,
high-reflectivity environments.

\section{Conclusion}

This paper presented a fully model-driven 4D radar perception framework for
robust human detection in harsh industrial and underground environments where
cameras and LiDAR often fail due to airborne particulates, low visibility, and
confined geometry. The proposed pipeline operates exclusively on 4D mmWave radar
point clouds and combines domain-aware multi-threshold filtering, two-frame
ego-motion–compensated accumulation, Doppler-aware clustering, and a lightweight
rule-based classifier. The system is designed for real-time execution on
embedded platforms and requires no supervised training.

Experiments in a controlled dust-filled trailer and a real underground mining
tunnel demonstrated that the 4D radar maintains a stable structure and reliable
pedestrian detection under severe visibility degradation, in conditions where
both vision-based YOLO detection and LiDAR returns deteriorate. These results
validate the suitability of elevation-resolved radar for pervasive safety
monitoring and human detection in environments dominated by dust, clutter, and
irregular reflectivity.

Overall, this work establishes a clear, interpretable, and deployable baseline
for radar-only perception in adverse environments, and serves as a foundation
for future 4D radar research in safety-critical industrial applications.

\section*{ACKNOWLEDGEMENT}
The authors of this paper would like to acknowledge the support of Rogers, LoopX, Purolator, Cloudhawk, NSERC, MITACS, and WATCAR.

\bibliographystyle{IEEEtran}
\bibliography{reference}

\end{document}